\documentclass[lettersize,journal]{IEEEtran}

\usepackage{changes}
\usepackage{amsmath,amsfonts}
\usepackage{multirow}
\usepackage{array}
\usepackage[caption=false,font=normalsize,labelfont=sf,textfont=sf]{subfig}
\usepackage{textcomp}
\usepackage{stfloats}
\usepackage{url}
\usepackage{verbatim}
\usepackage{graphicx}
\def\BibTeX{{\rm B\kern-.05em{\sc i\kern-.025em b}\kern-.08em
		T\kern-.1667em\lower.7ex\hbox{E}\kern-.125emX}}

\usepackage[ruled,linesnumbered]{algorithm2e}

\usepackage{hyperref}

\usepackage{xcolor}

\hypersetup{
	colorlinks=true,
	linkcolor=blue,
	citecolor=blue,
}

\usepackage{balance}

\usepackage{soul}

\usepackage[switch]{lineno}

\begin{document}
	
\title{An Adaptive ICP LiDAR Odometry Based on Reliable Initial Pose}
\author{
Qifeng Wang, \emph{Student Member, IEEE}, Weigang Li, \emph{Member, IEEE}, Lei Nie, \emph{Member, IEEE},  Xin Xu, \emph{Senior Member, IEEE}, Wenping Liu, \emph{Senior Member, IEEE}, Zhe Xu, \emph{Member, IEEE}
	
\IEEEcompsocitemizethanks{	
	\IEEEcompsocthanksitem This work was partially supported by the National Natural Science Foundation of China under Grant No. 51774219, the Key R\&D Program of Hubei Province under Grant No. 2020BAB098, and the Hubei Science and Technology Talents Service Enterprise Project under Grant No. 202400288. Numerical calculations were supported by the High-Performance Computing Center of Wuhan University of Science and Technology. \emph{(Corresponding author: Weigang Li.)}
	\IEEEcompsocthanksitem		Qifeng Wang, Weigang Li and Zhe Xu are with the School of Information Science and Engineering, Wuhan University of Science and Technology, Wuhan, China.
	\IEEEcompsocthanksitem		Lei Nie and Xin Xu are with the School of Computer Science and Technology, Wuhan University of Science and Technology, Wuhan, China. 
	\IEEEcompsocthanksitem		Wenping Liu is with the School of Information Management and Institute of Big Data and Digital Economy, Hubei University of Economics, Wuhan, China.
}}

\markboth{IEEE Transactions on Instrumentation and Measurement}{}
\maketitle
\begin{abstract}
As a key technology for autonomous navigation and positioning in mobile robots, light detection and ranging (LiDAR) odometry is widely used in autonomous driving applications. The Iterative Closest Point (ICP)-based methods have become the core technique in LiDAR odometry due to their efficient and accurate point cloud registration capability. However, some existing ICP-based methods do not consider the reliability of the initial pose, which may cause the method to converge to a local optimum. Furthermore, the absence of an adaptive mechanism hinders the effective handling of complex dynamic environments, resulting in a significant degradation of registration accuracy. To address these issues, this paper proposes an adaptive ICP-based LiDAR odometry method that relies on a reliable initial pose. First, distributed coarse registration based on density filtering is employed to obtain the initial pose estimation. The reliable initial pose is then selected by comparing it with the motion prediction pose, reducing the initial error between the source and target point clouds. Subsequently, by combining the current and historical errors, the adaptive threshold is dynamically adjusted to accommodate the real-time changes in the dynamic environment. Finally, based on the reliable initial pose and the adaptive threshold, point-to-plane adaptive ICP registration is performed from the current frame to the local map, achieving high-precision alignment of the source and target point clouds. Extensive experiments on the public KITTI dataset demonstrate that the proposed method outperforms existing approaches and significantly enhances the accuracy of LiDAR odometry.
\end{abstract}


\section{Introduction}
Light detection and ranging (LiDAR), with its high-precision distance measurement and 3D modeling capabilities, provides significant advantages in low-light environments. Consequently, it has become an essential sensor in autonomous driving and robotic systems \cite{10632209,zou2021comparative}. Leveraging its depth perception, LiDAR odometry technology has emerged to enhance autonomous navigation, enabling real-time motion state calculation and high-precision positioning in dynamic, complex environments  \cite{Light-LOAM,WANG2023112767,wen2023dynamic}.

Among the various LiDAR odometry methods, Iterative Closest Point (ICP)-based methods have become the mainstream due to their efficient and accurate point cloud registration capabilities \cite{icp,10474286,24}. The ICP-based methods estimate pose by iteratively aligning the source point cloud with the target point cloud through optimization. However, existing ICP-based methods present significant limitations in practical applications. Firstly, these methods often do not adequately consider the reliability of the initial pose estimation, making the algorithm susceptible to converging to local optima when the initial pose is inaccurate, thereby leading to registration failures or decreased accuracy \cite{LiTAMIN,9729241,fastergicp}. Secondly, traditional ICP-based methods lack adaptive mechanisms, making it challenging to effectively handle dynamic and complex environmental changes, such as numerous moving objects or rapidly changing conditions. This results in a substantial reduction in registration accuracy and negatively impacts the overall performance of LiDAR odometry \cite{PDLC-LIO,Fast-lio2,vizzo2023kiss}.

\begin{figure*}[htbp] 
	\centering 
	\includegraphics[width=0.7\linewidth]{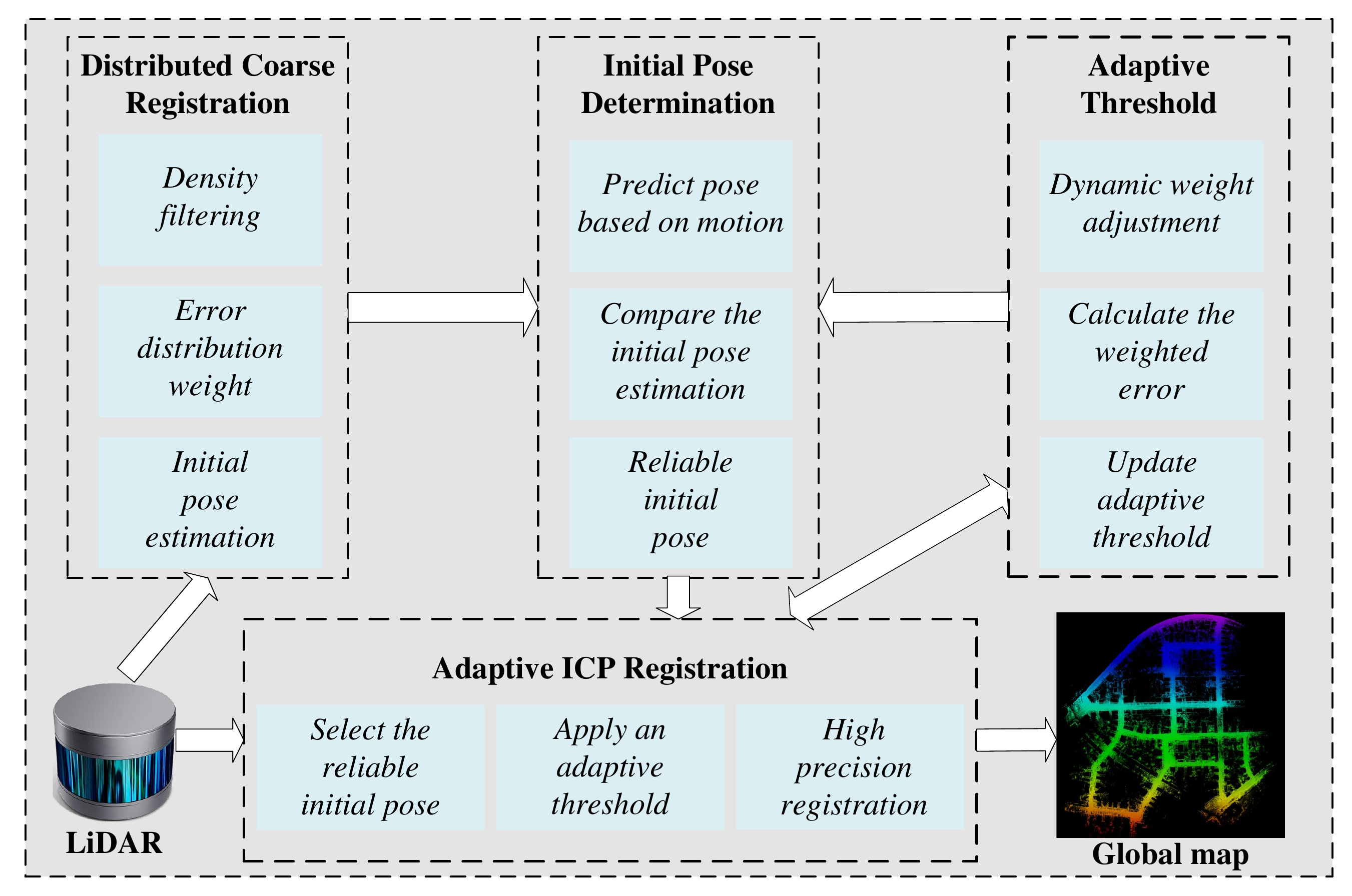}\\
	\caption{Overview of Our Approach includes four key components: distributed coarse registration, initial pose determination, adaptive threshold, and adaptive ICP registration.}
	\label{01}
\end{figure*}

To address these challenges, this paper proposes an adaptive ICP-based LiDAR odometry method based on a reliable initial pose. The method first preprocesses the source point cloud through distributed coarse registration using density filtering to obtain an initial pose estimation. Subsequently, historical pose information is incorporated for motion prediction and compared with the initial pose obtained in the previous step to select the most reliable initial pose, thereby reducing the initial error between the source and target point clouds. Furthermore, an adaptive threshold is dynamically adjusted based on the current and historical errors, which enables the method to flexibly adjust registration parameters according to different motion states and real-time environments. Finally, an adaptive weight mechanism is employed to weight each pair of points during the point cloud registration process, reducing the influence of outliers and achieving high-precision alignment of the source and target point clouds. Fig. \ref{01} shows the model of our approach. To verify the effectiveness of the proposed method, we conducted extensive experimental evaluations using the public KITTI dataset and compare our approach with the latest mainstream LiDAR odometry methods. The experimental results demonstrate that the proposed method outperforms existing approaches in registration accuracy, effectively enhancing the overall performance of LiDAR odometry in complex dynamic environments.

The main contributions of our work are as follows: 
\begin{itemize} 
	\item 
	To address the issue of unreliable initial poses leading to local optimal solutions, a method for obtaining a more reliable initial pose is proposed. Distributed coarse registration based on density filtering is first employed to estimate the initial pose. The reliable initial pose is then selected by comparing it with the motion prediction pose, thereby minimizing the initial error between the source and target point clouds.
	
	\item To solve the problem of insufficient adaptability in dynamic environment, an adaptive frame-to-local map ICP registration method is proposed. The registration parameters were dynamically adjusted by combining current and historical errors to enhance the adaptability of the method in dynamic environments.
	
	\item To evaluate the effectiveness of these methods in real-world scenarios, extensive experiments were conducted using the public KITTI dataset \cite{vgicp,fastergicp,DLO,vizzo2023kiss}. These experiments compared our method with existing mainstream LiDAR odometry methods. The results demonstr ate that the proposed method is superior to other mainstream odometry methods in terms of accuracy.
\end{itemize}

The rest of this paper is as follows: Section II introduces related work and reviews the current research progress in LiDAR odometry and the ICP-based methods. Section III elaborates on the theoretical foundation and implementation steps of the proposed method. Section IV presents the experimental results and analysis. Finally, Section V concludes the paper and discusses future research directions.

\section{Related Work}
LiDAR odometry, a crucial technology for autonomous navigation and localization in mobile robots, has seen widespread application in recent years across various fields, including autonomous driving, unmanned aerial vehicles, and intelligent robotics. Generally, LiDAR odometry methods are categorized into two main types: feature-based methods \cite{30,I-LOAM} and  ICP-based methods \cite{LiTAMIN2,igicp}.

\subsection{Feature-Based LiDAR Odometry} 
Feature-based LiDAR odometry methods achieve accurate localization and map construction by extracting geometric elements from the data, such as points, line segments, and planes, to establish associations between environmental features \cite{Liosam,39,Lego-loam,behley2018efficient}. The LOAM method, proposed by Dr. Zhang from Carnegie Mellon University, represents a significant advancement in this domain \cite{LOAM}. LOAM effectively balances computational efficiency and localization accuracy by integrating high-frequency, lower-precision odometry updates with low-frequency, higher-precision map matching. The method divides the motion estimation process into two complementary steps, thereby accelerating processing speed and reducing computational complexity. This makes LOAM particularly well-suited for applications that require both real-time performance and high accuracy. Subsequently, Shan et al. extended the LOAM method by developing LeGO-LOAM, which enhances feature extraction accuracy through the incorporation of ground segmentation and target clustering techniques \cite{Lego-loam}. Additionally, this method integrates a loop closure detection module, which effectively mitigates pose drift caused by accumulated errors. As a further optimization of LOAM, F-LOAM significantly enhances system operational efficiency while maintaining high-precision registration through improved feature extraction and data matching methods \cite{F-loam}. Feature-based LiDAR methods can effectively identify and track key geometric features in the environment, making them suitable for reliable localization and map generation in diverse complex scenarios. However, these methods perform well in feature-rich environments but often struggle with feature extraction in structurally simple or feature-scarce environments, leading to reduced registration accuracy \cite{fastergicp,fastlio}.

In recent years, deep learning techniques have been applied to LiDAR odometry. Deep learning-based feature methods can automatically learn useful features from large datasets, reducing the complexity of manual feature design. Li et al. proposed LO-Net, a deep convolutional network that utilizes a weighted geometric constraint loss to efficiently learn features for LiDAR odometry estimation \cite{li2019net}. Yuan et al. introduced SDV-LOAM, which applies deep learning to optimize the LOAM method in complex environments, improving localization and mapping accuracy \cite{yuan2023sdv}. However, deep learning methods require large training datasets and considerable computational resources, which makes the training process time-consuming and expensive, posing a challenge to real-time odometry requirements. Additionally, the black-box nature of neural networks renders their decision-making process difficult to interpret \cite{PDLC-LIO,Dlc-slam,igicp}.

\subsection{ICP-Based LiDAR Odometry} 
Compared to feature-based methods, ICP-based methods directly process point cloud data without relying on feature extraction, making them superior in environments where features are sparse or feature extraction is suboptimal \cite{FAST-LIVO,Lion,loam22}. As a classic point cloud alignment method, ICP-based methods iteratively compute the minimum distance between two sets of point clouds to determine the optimal rotation and translation, thereby achieving precise alignment of the two datasets. Building on this, GICP introduces a Gaussian probability model, incorporating uncertainty parameters into the ICP cost function to improve computational efficiency and real-time performance \cite{Gicp}. Fast-gicp optimizes data processing by converting point cloud data into a voxel grid structure and performing ICP registration on this basis, significantly reducing computational complexity and meeting the high-efficiency processing needs of large-scale point clouds \cite{vgicp}. Furthermore, Faster-gicp enhances GICP by using acceptance-rejection sampling methods to filter and associate point cloud data, effectively reducing the number of data points to be processed and significantly increasing both the computational speed and matching accuracy of the method \cite{fastergicp}. LITAMIN2 improves registration speed and accuracy by approximating the distribution of local geometries with normal distributions, making it particularly suitable for processing high-density LiDAR data \cite{LiTAMIN2}. DLO achieves real-time high-precision pose estimation by utilizing dense and minimally preprocessed point cloud data, making it ideal for long-term operation in robotic systems and effectively enhancing system accuracy and stability \cite{DLO}. The Kiss-icp method, based on the classical ICP method, employs multi-level filtering and robust feature matching strategies to ensure stable localization performance even in complex environments \cite{vizzo2023kiss}.

Although ICP-based methods have significantly reduced errors and enhanced system robustness in the field of LiDAR odometry, which have become the preferred technology in many application scenarios, they still exhibit notable limitations in practical applications \cite{LiTAMIN,igicp}. Firstly, these methods often do not fully consider the reliability of initial pose estimates, causing the method to easily converge to local optima and resulting in registration failure or decreased accuracy when the initial pose is inaccurate. Secondly, these methods lack adaptive mechanisms, making it difficult to effectively handle dynamic and complex environmental changes, such as numerous moving objects or rapidly changing conditions, which significantly reduces registration accuracy and affects the overall performance of LiDAR odometry \cite{PDLC-LIO,9765591,Fast-lio2}.

To overcome these issues, this paper proposes an adaptive ICP-based LiDAR odometry method, which relies on reliable initial pose estimation, and solves the problem of inadequate adaptability of existing ICP-based methods in dynamic environments while maintaining high-precision registration accuracy.
\section{Our Approach}
To enhance the registration accuracy of LiDAR odometry in dynamic environments, this paper proposes an adaptive ICP-based LiDAR odometry method. As illustrated in Fig. \ref{01}, the method consists of the following key steps: First, distributed coarse registration, based on density filtering, is employed to obtain initial pose estimates, thereby enhancing their reliability. Subsequently, the initial pose from coarse registration is compared with the pose derived from motion prediction, and the more reliable initial pose is selected to minimize the initial error between the source and target point clouds. Next, by combining the current frame's registration error with historical error data, the adaptive threshold is dynamically adjusted. This adjustment allows the registration method to flexibly modify optimization parameters in response to real-time environmental changes, thus effectively addressing challenges in dynamic environments. Finally, based on the reliable initial pose and adaptive threshold, a Point-to-Plane ICP registration strategy is employed to achieve high-precision alignment between the current frame and the local map, ensuring accurate registration of the source and target point clouds.

\subsection{Distributed Coarse Registration}
In point cloud registration, providing a reliable and high-quality initial pose is crucial for ensuring registration accuracy and robustness. However, source and target point clouds often contain noise and irregular sparse data, which may significantly degrade the registration results. To address these issues, this paper presents a distributed coarse registration method that utilizes density filtering. {The flow of the distributed coarse registration is summarized in Algorithm \ref{coarse_registration}.}

\begin{algorithm}[htbp]
	\small
	\setlength{\abovedisplayskip}{2pt}
	\setlength{\belowdisplayskip}{2pt}
	\caption{Distributed Coarse Registration}
	\label{coarse_registration}
	\KwIn{Source cloud $\mathbf{S}$, Target cloud $\mathbf{S}_t$, Radius $r$, Neighborhood size $k$, Density threshold percentile $\alpha$}
	\KwOut{Initial transformation matrix $\mathbf{T}_{\mathrm{align},i}$}
	
	\For{each point $\mathbf{p}_m \in \mathbf{S}$}{
		Compute density:
		\begin{align*}
			D_m = \sum_{n=1}^{M} \mathbb{I}(\| \mathbf{p}_m - \mathbf{p}_n \| \leq r)
		\end{align*}
	}
	
	Compute density threshold:
	\begin{align*}
		D_{\mathrm{th}} = Q_{\alpha}(D)
	\end{align*}
	
	Filter $S$: retain points where $D_m \geq D_{\mathrm{th}}$\;
	
	\For{each point $\mathbf{p}_m \in \mathbf{S}$}{
		Compute covariance matrix:
		\begin{align*}
			\mathbf{C}_m = \frac{1}{k} \sum_{n=1}^{k} (\mathbf{p}_n - \mathbf{\mu}_m)(\mathbf{p}_n - \mathbf{\mu}_m)^\top
		\end{align*}
		where $\mathbf{\mu}_m = \frac{1}{k} \sum_{n=1}^{k} \mathbf{p}_n$
	}
	
	\For{each point $\mathbf{p}_m \in \mathbf{S}$}{
		Transform point:
		\begin{align*}
			\mathbf{p}'_m = \mathbf{T} \mathbf{p}_m
		\end{align*}
		
		Find closest point in target cloud:
		\begin{align*}
			\mathbf{q}_n = \arg\min_{\mathbf{q} \in \mathbf{S}_t} \|\mathbf{p}'_m - \mathbf{q}\|^2
		\end{align*}
		
		Compute error vector:
		\begin{align*}
			\mathbf{e}_m = \mathbf{q}_n - \mathbf{p}'_m
		\end{align*}
		
		Compute joint covariance matrix:
		\begin{align*}
			\mathbf{M}_m = \mathbf{C}_m + \mathbf{C}_n
		\end{align*}
		
		Compute weight:
		\begin{align*}
			w_m = \exp\left(-\frac{\mathbf{e}_m^\top \mathbf{M}_m^{-1} \mathbf{e}_m}{2\sigma^2}\right)
		\end{align*}
	}
	
	Solve for transformation update:
	\begin{align*}
		\mathbf{H} = \sum_{m} w_m \mathbf{J}_m^\top \mathbf{M}_m^{-1} \mathbf{J}_m, \quad
		\mathbf{b} = \sum_{m} w_m \mathbf{J}_m^\top \mathbf{M}_m^{-1} \mathbf{e}_m
	\end{align*}
	
	Compute $\Delta \mathbf{x}$ and update transformation:
	\begin{align*}
		\mathbf{T}_{\mathrm{align},i} = \exp\left( (\mathbf{H} + \lambda \mathbf{I})^{-1} \mathbf{b} \right) \mathbf{T}
	\end{align*}
	
	\Return $\mathbf{T}_{\mathrm{align},i}$
\end{algorithm}

First, for each point $\mathbf{p}_m$ in the source point cloud $\mathbf{S}$, the number of neighboring points $D_m$ within a radius $r$ is calculated.
\begin{equation}
	D_m = \sum_{n=1}^{M} \mathbb{I}\left( \| \mathbf{p}_m - \mathbf{p}_n \| \leq r \right) ,
\end{equation}
where $M$ is the number of neighboring points and  $\mathbb{I}(\cdot)$ is the indicator function, returning 1 if point $\mathbf{p}_n$ is within radius $r$ of point $\mathbf{p}_m$, and 0 otherwise.

By computing the density distribution $D = \{ D_1, D_2, \ldots, D_t \}$ of all points in the source point cloud and setting a density threshold $D_{\mathrm{th}} = Q_{\alpha}(D)$, where $Q_{\alpha}(D)$ is the $\alpha$-percentile of the density distribution, only points with density $D_m \geq D_{\mathrm{th}}$ are retained for subsequent registration steps. 

To enhance geometric consistency during the registration process, we compute the local surface covariance matrix for each point. For each point $\mathbf{p}_m$ in the source point cloud, the nearest $k$ neighboring points are identified using a $k$-d tree, and the covariance matrix $\mathbf{C}_m$ is calculated as follows.
\begin{equation}
	\mathbf{C}_m = \frac{1}{k} \sum_{n=1}^{k} (\mathbf{p}_n - \mathbf{\mu}_m)(\mathbf{p}_n - \mathbf{\mu}_m)^\top ,
\end{equation}
where $\mathbf{\mu}_m$ is the mean vector of point $\mathbf{p}_m$ and its neighboring points.
\begin{equation}
	\mathbf{\mu}_m = \frac{1}{k} \sum_{n=1}^{k} \mathbf{p}_n  .
\end{equation}

After completing  covariance calculation for the point cloud, we utilize the covariance matrices to describe the local distribution characteristics of the points, achieving more accurate correspondences. Assume the current transformation matrix is $\mathbf{T}$. Each point $\mathbf{p}_m$ in the source point cloud is transformed to $\mathbf{p}'_m = \mathbf{T} \mathbf{p}_m$, and then the closest point $\mathbf{q}_n$ in the target point cloud is searched, where $\mathbf{S}_t$ denotes the set of points in the target point cloud.
\begin{equation}
	\mathbf{q}_n = \arg\min_{\mathbf{q} \in \mathbf{S}_t} \| \mathbf{p}'_m - \mathbf{q} \|^2 .
\end{equation}

For each valid pair of corresponding points, the geometric error vector $\mathbf{e}_m = \mathbf{q}_n - \mathbf{p}'_m$ is calculated and weighted using the covariance matrices of the source and target points. Specifically, the joint covariance matrix $\mathbf{M}_m = \mathbf{C}_{m} + \mathbf{C}_{n}$ is constructed using the covariance matrices $\mathbf{C}_{m}$ and $\mathbf{C}_{n}$ of the $\mathbf{p}_m$ and $\mathbf{q}_n$, respectively. The error vector is then weighted using $\mathbf{M}_m$, resulting in the weighted squared norm of the error $d_{mn}^2$.
\begin{equation}
	d_{mn}^2 = \mathbf{e}_m^\top \mathbf{M}_m^{-1} \mathbf{e}_m .
\end{equation}

We define the error distribution weight $w_m$.
\begin{equation}
	w_m = \exp\left( -\frac{d_{mn}^2}{2\sigma^2} \right) ,
\end{equation}
where $\sigma$ is a tuning parameter that controls the rate of weight decay.

Next, construct the weighted Hessian matrix $\mathbf{H}$ and the gradient vector $\mathbf{b}$.
\begin{equation}
	\mathbf{H} = \sum_{m} w_m \mathbf{J}_m^\top \mathbf{M}_m^{-1} \mathbf{J}_m ,
\end{equation}
\begin{equation}
	\mathbf{b} = \sum_{m} w_m \mathbf{J}_m^\top \mathbf{M}_m^{-1} \mathbf{e}_m ,
\end{equation}
here, $\mathbf{J}_m$ is the Jacobian matrix of the error vector $\mathbf{e}_m$ with respect to the transformation parameters. 

Then, compute the transformation increment $\Delta \mathbf{x}$.
\begin{equation}
	\Delta \mathbf{x} = (\mathbf{H} + \lambda \mathbf{I})^{-1} \mathbf{b} ,
	\label{eq09}
\end{equation}
where $\lambda$ is the damping factor used to adjust the stability and convergence speed of the optimization process, and $\mathbf{I}$ is the identity matrix. Finally, update the transformation matrix $\mathbf{T}_{\mathrm{align},i} = \exp(\Delta \mathbf{x}) \mathbf{T}$, where $i$ denotes the $i$-th frame, and $\exp(\Delta \mathbf{x})$ represents the application of the incremental transformation $\Delta \mathbf{x}$ to the current transformation matrix $\mathbf{T}$ via the Lie group exponential map.

By weighting the errors using the local covariance matrices of the source and target points, the process converges to the optimal transformation matrix $\mathbf{T}_{\mathrm{align},i}$. This provides a high-precision initial pose estimate for subsequent fine registration.
\subsection{Initial Pose Determination}
The accuracy of the initial pose is crucial for the convergence rate of the registration process and the precision of the final results. By combining the initial pose estimates from distributed coarse registration using density filtering with historical pose information, we design a novel initial pose determination method.

We generate a predicted transformation matrix \( \mathbf{T}_{\mathrm{pred},i} \) by combining historical pose information and current point cloud features. The set of historical poses is defined as \( \{\mathbf{T}_1, \mathbf{T}_2, \ldots, \mathbf{T}_i\} \), where \( i \) is the number of historical poses. Depending on the number of historical poses, different processing strategies are adopted.

When \( i < 2 \), there is insufficient historical pose information for effective prediction, so we return the identity transformation matrix \( \mathbf{I} \) as the predicted transformation matrix.

When \( i = 2 \), we directly compute the relative transformation between the two historical poses as \( \mathbf{T}_{\mathrm{pred},i} = \mathbf{T}_{i-1}^{-1} \cdot \mathbf{T}_i \).

When \( i \geq 3 \), we select the most recent three historical poses \( \{\mathbf{T}_{i-3}, \mathbf{T}_{i-2}, \mathbf{T}_{i-1}\} \) and calculate the incremental transformations \( \Delta \mathbf{T}_j = \mathbf{T}_j^{-1} \cdot \mathbf{T}_{j+1} \) where \( j = i-3, i-2 \).

Subsequently, we introduce an adaptive threshold \( \sigma_{\mathrm{th}} \) (in Section C) to adjust the weights \( w_{r,i} \) and \( w_{e,i} \) of the two incremental transformations \( \Delta \mathbf{T}_{i-3} \) and \( \Delta \mathbf{T}_{i-2} \), where \( w_{r,i} \) is positively correlated with \( \sigma_{\mathrm{th}} \). This ensures that when the error is large, the system relies more on the most recent pose transformations, thereby generating the predicted transformation matrix.
\begin{equation}
	\mathbf{T}_{\mathrm{pred},i} = \frac{w_{e,i} \cdot \Delta \mathbf{T}_{i-3} + w_{r,i} \cdot \Delta \mathbf{T}_{i-2}}{w_{e,i} + w_{r,i}} \cdot \mathbf{T}_{i-1} .
\end{equation}

Next, by comparing the translation difference \( \Delta t_i \) between \( \mathbf{T}_{\mathrm{pred},i} \) and the pose \( \mathbf{T}_{\mathrm{align},i} \) generated by distributed coarse registration based on density filtering, we determine the final reliable initial pose \( \mathbf{T}_{\mathrm{init},i} \).
\begin{equation}
	\Delta t_i = \| (\mathbf{T}_{\mathrm{pred},i}^{-1} \cdot \mathbf{T}_{\mathrm{align},i})_{\mathrm{trans}} \| , 
\end{equation}
here, \((\cdot)_{\mathrm{trans}}\) denotes the translation component of the transformation matrix.

Finally, based on the translation difference \( \Delta t_i \) and a predefined translation difference threshold \( \tau \), we select the final initial pose \( \mathbf{T}_{\mathrm{init},i} \). When \( \Delta t_i \) does not exceed the threshold \( \tau \), \( \mathbf{T}_{\mathrm{align},i} \) is chosen as the final initial pose. Otherwise, the motion-predicted pose \( \mathbf{T}_{\mathrm{pred},i} \) is selected, to prevent \( \mathbf{T}_{\mathrm{align},i} \) from falling into a local optimum and resulting in large errors, which would lead to inaccurate subsequent registration processes. This method not only optimizes the registration process but also ensures the accuracy of our approach in complex environments.

\subsection{Adaptive Threshold}
During the registration process, the system's motion state may vary over time, requiring the registration method to adjust to dynamic environments to maintain high precision and robustness. Therefore, we propose an adaptive threshold method that dynamically adjusts the threshold by combining the current error with historical error information, thereby adapting to varying motion states and changes in the point cloud environment.

First, the system obtains the acceleration \( \mathbf{\dot{v}}_i \) of the current frame relative to the previous frame. Then, the difference between the current acceleration \( \mathbf{\dot{v}}_i \) and the acceleration of the previous frame \( \mathbf{\dot{v}}_{i-1} \) is calculated to obtain the acceleration change rate \( \alpha_i = \| \mathbf{\dot{v}}_i - \mathbf{\dot{v}}_{i-1} \| \). Based on \( \alpha_i \), an exponential decay function is used to compute the dynamic weight \( \gamma_i \), reflecting the stability of the motion state.
\begin{equation}
	\gamma_i = \exp\left( -\frac{\alpha_i}{\Delta t \cdot \sigma_{\mathrm{decay}}} \right) \label{eq:dynamic_weight} ,
\end{equation}
where \( \Delta t \) is the time interval and \( \sigma_{\mathrm{decay}} \) is the decay factor.

After computing \( \gamma_i \), the model deviation matrix \( \mathbf{D}_{\mathrm{model},i} \) between  \( \mathbf{T}_{\mathrm{init},i} \) and \( \mathbf{T}_{\mathrm{new},i} \) is calculated.
\begin{equation}
	\mathbf{D}_{\mathrm{model},i} =
	\mathbf{T}_{\mathrm{init},i}^{-1} \cdot \mathbf{T}_{\mathrm{new},i} =
	\begin{bmatrix}
		\mathbf{R}_i & \mathbf{t}_i \\
		0 & 1
	\end{bmatrix} \label{eq:model_deviation} ,
\end{equation}
where \( \mathbf{R}_i \) is the rotation matrix representing the rotational component, and \( \mathbf{t}_i \) is the translation vector representing the translational component.

Next, the weighted error \( e_i \) is defined as the weighted sum of the rotation error and translation error.
\begin{equation}
	e_i = \sigma_{\mathrm{max}} \cdot \tanh(\beta \cdot \theta_i) + \|\mathbf{t}_i\| ,
\end{equation}
where \( \sigma_{\mathrm{max}} \) is a preset maximum rotation error scaling factor to limit the maximum value of the rotation error, thereby preventing excessive influence of the rotation error on the overall error. \( \beta \) is a preset scaling factor used to control the nonlinear scaling degree of the rotation error, and \( \theta_i \) is the rotation angle of the \( i \)-th frame, calculated from \( \mathbf{R}_i \).
\begin{equation}
	\theta_i = \cos^{-1}\left(\frac{\mathrm{tr}(\mathbf{R}_i) - 1}{2}\right) .
\end{equation}

Finally, the adaptive threshold \( \sigma_{\mathrm{th}} \) is calculated using the root mean square of the errors.
\begin{equation}
	\sigma_{\mathrm{th}} = \sqrt{\frac{\sum_{i=1}^{N} \gamma_i \cdot e_i^2}{N}} ,
\end{equation}
where \( N \) is the number of frames. 

This method ensures that the threshold can be flexibly adjusted based on dynamic motion states and historical error information, significantly enhancing the accuracy of the ICP registration process.

\subsection{Adaptive ICP Registration}
After determining \( \mathbf{T}_{\mathrm{init},i} \) and \( \sigma_{\mathrm{th}} \), adaptive ICP registration is performed to obtain a new pose estimate. {The flow of the adaptive ICP registration is summarized in Algorithm \ref{adaptive_icp}.}

\begin{algorithm}[htbp]
	\small
	\setlength{\abovedisplayskip}{2pt}
	\setlength{\belowdisplayskip}{2pt}
	\caption{Adaptive ICP Registration}
	\label{adaptive_icp}
	\KwIn{Source point cloud $\mathbf{S}$, Target point cloud $\mathbf{S}_t$, Initial transformation $\mathbf{T}_{\mathrm{init},i}$, Adaptive threshold $\sigma_{\mathrm{th}}$, Maximum iterations $N_{\mathrm{max}}$}
	\KwOut{Optimized transformation $\mathbf{T}_{\mathrm{new},i}$}
	
	Initialize transformation: $\mathbf{T}_{\mathrm{new},i} = \mathbf{T}_{\mathrm{init},i}$\;
	
	\For{$i = 1$ to $N_{\mathrm{max}}$}{
		Transform source point cloud:
		\[
		\mathbf{S}' = \mathbf{T}_{\mathrm{new},i} \cdot \mathbf{S}
		\]
		
		\For{each point $\mathbf{s}_k \in \mathbf{S}'$}{
			Find nearest neighbor $\mathbf{q}_k \in \mathbf{S}_t$\;
			Compute point-to-plane residual:
			\[
			e_k = (\mathbf{p}_k - \mathbf{q}_k) \cdot \mathbf{n}_k
			\]
			
			Compute adaptive weight:
			\[
			\beta_k = \frac{\sigma_{\mathrm{th}}^2}{\sigma_{\mathrm{th}}^2 + e_k^2}
			\]
		}
		
		Construct Jacobian matrix:
		\[
		\mathbf{J}_k = \begin{bmatrix}
			\mathbf{S}(\mathbf{p}_k) & -\mathbf{I} 
		\end{bmatrix}
		\]
		
		Compute Hessian matrix and gradient:
		\[
		\mathbf{H} = \sum_{k=1}^{M} \beta_k \mathbf{J}_k^\top \mathbf{J}_k, \quad
		\mathbf{b} = \sum_{k=1}^{M} \beta_k \mathbf{J}_k^\top \mathbf{e}_k
		\]
		
		Compute transformation increment:
		\[
		\Delta \mathbf{x} = (\mathbf{H} + \lambda \mathbf{I})^{-1} \mathbf{b}
		\]
		
		Update transformation:
		\[
		\mathbf{T}_{\mathrm{new},i} = \exp(\Delta \mathbf{x}) \cdot \mathbf{T}_{\mathrm{new},i}
		\]
		
 		\If{$\Delta \mathbf{x}$ update is negligible}{break\;}
	}
	
	\Return $\mathbf{T}_{\mathrm{new},i}$
\end{algorithm}

First, apply \( \mathbf{T}_{\mathrm{init},i} \) to the source point cloud \( \mathbf{S} \) to obtain the transformed point cloud \( \mathbf{S}' \).
\begin{equation}
	\mathbf{S}' = \mathbf{T}_{\mathrm{init},i} \cdot \mathbf{S} ,
\end{equation}
the purpose of this transformation is to bring the source point cloud as close as possible to the target point cloud under the initial estimate.

Next, for each transformed source point \( \mathbf{s}_k\), find its nearest neighboring target point \( \mathbf{q}_k \) in the target point cloud  $\mathbf{S}_t$. To measure the alignment error between the source and target points, calculate the point-to-plane residual \( e_k \) using the normal vector \( \mathbf{n}_k \) of the target point.
\begin{equation}
	e_k = (\mathbf{p}_k - \mathbf{q}_k) \cdot \mathbf{n}_k ,
\end{equation}
where \( \mathbf{p}_k = \mathbf{T}_{\mathrm{init},i} \cdot \mathbf{s}_k \) represents the position of the transformed source point.

To further optimize the pose estimation, the system constructs the Jacobian matrix \( \mathbf{J}_k \), which describes the linearized relationship between the residual vector and the pose parameters.
\begin{equation}
	\mathbf{J}_k = \begin{bmatrix}
		\mathbf{S}(\mathbf{p}_k) & -\mathbf{I} 
	\end{bmatrix} ,
\end{equation}
where \( \mathbf{S}(\mathbf{p}_k) \) is the skew-symmetric matrix of the vector \( \mathbf{p}_k \), used to map changes in the rotational part to the residual vector, effectively capturing the influence of pose parameters on the error. \( \mathbf{I} \) is the \( 3 \times 3 \) identity matrix, corresponding to the translational part.
\begin{equation}
	\mathbf{S}(\mathbf{p}_k) = \begin{bmatrix}
		0 & -p_{k,z} & p_{k,y} \\
		p_{k,z} & 0 & -p_{k,x} \\
		-p_{k,y} & p_{k,x} & 0 
	\end{bmatrix} .
\end{equation}

To enhance the method's robustness against noise and outliers, an adaptive mechanism is introduced by dynamically adjusting the weight of each residual. When the residual is small, the weight \( \beta_k \) approaches 1, maintaining its primary contribution to the optimization; when the residual is large, \( \beta_k \) rapidly decreases, thereby diminishing the influence of outlier points on the optimization.
\begin{equation}
	\beta_k = \frac{\sigma_{\mathrm{th}}^2}{\sigma_{\mathrm{th}}^2 + e_k^2} .
\end{equation}

By accumulating the weighted residual vectors and Jacobian matrices of all valid point pairs, a weighted linear system is constructed to minimize the sum of the weighted squared residuals.
\begin{equation}
	\mathbf{H} = \sum_{k=1}^{M} \beta_k \mathbf{J}_k^\top \mathbf{J}_k, \quad \mathbf{b} = \sum_{k=1}^{M} \beta_k \mathbf{J}_k^\top \mathbf{e}_k ,
\end{equation}
where \( M \) is the number of valid point pairs.

The transformation increment \( \Delta \mathbf{x} \) is then used to update the current pose estimate. Consistent with Equation (\ref{eq09}), the Levenberg-Marquardt method is employed to compute the transformation increment \( \Delta \mathbf{x} \). The pose estimate update \( \Delta \mathbf{x} \) is mapped back to the Lie algebra and multiplied with the initial pose to achieve the latest pose estimate.
\begin{equation}
	\mathbf{T}_{\mathrm{new},i} = \exp(\Delta \mathbf{x}) \cdot \mathbf{T}_{\mathrm{init},i} .
\end{equation}

By utilizing reliable initial pose and introducing adaptive threshold, the geometric error between the source and target point clouds is effectively reduced, gradually refining the pose estimation to achieve accurate alignment. This process not only improves registration accuracy but also increases the method's robustness against noise and outliers, ensuring the reliability and stability of the registration process.
\section{Experiments}
{To verify the validity and accuracy of the proposed method, we conducted comparative experiments on public KITTI and MDGR datasets, comparing our method with current state-of-the-art lidar odometer technology. Additionally, we performed ablation studies to validate the contributions of distributed coarse registration based on density filtering and adaptive ICP registration from point cloud frames to the local map.	
For implementation, we utilized PCL 1.12 for point cloud processing, Eigen 3.4 for matrix operations, and Ceres Solver 2.1 for non-linear optimization. The system was developed on Ubuntu 18.04 with ROS Melodic, using C++14 as the programming standard.
To ensure a fair comparison, motion compensation for the point cloud was consistently applied across all baseline methods. This guarantees that all methods operate under the same conditions without bias introduced by motion distortion. All experiments were executed on a computer equipped with an Intel i7-13700F CPU and 32GB of memory.}

\subsection{Accuracy Evaluation and Comparison}
The KITTI Odometry dataset provides ground truth trajectories for 11 sequences, encompassing urban roads (sequences 00, 06, 07, 08), rural roads (sequences 02, 03, 04, 05, 09, 10), and highways (sequence 01). These diverse dynamic environments offer comprehensive benchmarks for evaluating the robustness and accuracy of odometry methods.

To assess the performance of our method, we compared it with leading LiDAR odometry approaches: Fast-gicp \cite{vgicp}, Faster-gicp \cite{fastergicp}, DLO \cite{DLO}, and Kiss-icp \cite{vizzo2023kiss}. Absolute Pose Error (APE) measures the direct difference between the estimated pose and the ground truth pose, providing a clear indication of method accuracy and global trajectory consistency. We utilized Root Mean Square Error (RMSE), Mean, and Standard Deviation (Std) of APE as evaluation metrics for method accuracy, as presented in Table \ref{tab1}.
\begin{table*}[]
	\centering
		\caption{{APE comparison of KITTI (METER)}\label{tab1}}
	\resizebox{0.9\linewidth}{!}{
		\begin{tabular}{c|ccc|ccc|ccc|ccc|ccc}
			\hline
			\multirow{3}{*}{Sequence} & \multicolumn{3}{c|}{\multirow{2}{*}{Fast-gicp \cite{vgicp}}} & \multicolumn{3}{c|}{\multirow{2}{*}{Faster-gicp \cite{fastergicp}}} & \multicolumn{3}{c|}{\multirow{2}{*}{DLO \cite{DLO}}} & \multicolumn{3}{c|}{\multirow{2}{*}{Kiss-icp \cite{vizzo2023kiss}}} & \multicolumn{3}{c}{\multirow{2}{*}{Ours}}     \\
			& \multicolumn{3}{c|}{}                           & \multicolumn{3}{c|}{}                             & \multicolumn{3}{c|}{}                     & \multicolumn{3}{c|}{}                          & \multicolumn{3}{c}{}                          \\ \cline{2-16} 
			& RMSE           & Mean           & Std           & RMSE            & Mean           & Std            & RMSE      & Mean      & Std               & RMSE              & Mean             & Std     & RMSE          & Mean          & Std           \\ \hline
			00                         & 15.15          & 5.82           & 3.81          & 10.14           & 8.24           & 5.91           & 3.28      & 3.08      & \textbf{1.13}     & 4.04              & 3.51             & 2.01    & \textbf{3.23} & \textbf{2.87} & 1.48          \\
			01                         & 23.67          & 22.20           & 10.53         & 24.21           & 21.87          & 10.39          & 120.10     & 113.8     & 38.40              & 20.18             & 18.16            & 8.80    & \textbf{15.30} & \textbf{13.70} & \textbf{6.77} \\
			02                         & 17.70           & 16.43          & 6.58          & 15.24           & 13.79          & 6.51           & 7.54      & 7.05      & 2.67              & 7.67              & 7.22             & 2.57    & \textbf{6.07} & \textbf{5.66} & \textbf{2.17} \\
			03                         & 2.74           & 2.51           & 1.10           & 1.91            & 1.73           & 0.80            & 1.09      & 1.02      & 0.39              & \textbf{0.84}     & \textbf{0.78}    & 0.33    & \textbf{0.84} & \textbf{0.78} & \textbf{0.29} \\
			04                         & 0.93           & 0.88           & 0.31          & 0.53            & 0.47           & 0.24           & 0.59      & 0.52      & 0.28              & 0.46              & 0.42             & 0.19    & \textbf{0.26} & \textbf{0.23} & \textbf{0.10} \\
			05                         & 3.90           & 3.54           & 1.64          & 4.04            & 3.46           & 2.08           & 1.99      & 1.80      & 0.84              & 2.07              & 1.87             & 0.89    & \textbf{1.30} & \textbf{1.23} & \textbf{0.42} \\
			06                         & 1.42           & 1.34           & 0.48          & 1.27            & 1.17           & 0.49           & 1.03      & 0.94      & 0.43              & 0.93              & 0.87             & 0.34    & \textbf{0.80} & \textbf{0.73} & \textbf{0.32} \\
			07                         & 0.76           & 0.65           & 0.40          & 1.04            & 0.88           & 0.55           & 0.90      & 0.76      & 0.49              & 0.44              & 0.40             & 0.19    & \textbf{0.43} & \textbf{0.39} & \textbf{0.17} \\
			08                         & 6.46           & 5.40           & 3.54          & 6.40             & 5.68           & 2.93           & 4.09      & 3.67      & 1.81              & 4.01              & 3.49             & 2.04    & \textbf{3.23} & \textbf{2.79} & \textbf{1.62} \\
			09                         & 2.26           & 1.86           & 1.28          & 2.77            & 2.15           & 1.74           & 2.69      & 2.44      & 1.12              & 1.97              & 1.72             & 0.96    & \textbf{1.16} & \textbf{1.04} & \textbf{0.51} \\
			10                         & 2.05           & 1.88           & 0.82          & 2.29            & 2.06           & 0.99           & 2.59      & 2.25      & 1.28              & 1.78              & 1.58             & 0.82    & \textbf{1.32} & \textbf{1.18} & \textbf{0.58} \\
			Avg                         & 7.00          & 5.68          & 2.77        & 6.35           & 5.59          & 2.97         & 13.26     & 12.49     & 4.44              & 4.04             & 3.64            & 1.74    & \textbf{3.08} & \textbf{2.78} & \textbf{1.31} \\ \hline
		\end{tabular}
	}
\end{table*}

From Table \ref{tab1}, it is evident that our method consistently outperforms the other comparison methods in terms of both RMSE and Mean across most sequences. For example, in sequence 00, our method achieves an RMSE of 3.23, which is significantly lower than Fast-gicp's 15.15, Faster-gicp's 10.14, and Kiss-icp's 4.04. In sequence 01, our method achieves an RMSE of 15.30, which is notably lower than DLO's 120.10, resulting in an RMSE reduction of approximately 87\%. This demonstrates the effectiveness of our method in improving accuracy, especially in challenging sequences. Similarly, in sequences 02 and 03, our method attains RMSEs of 6.07 and 0.84, respectively, further highlighting its robustness and consistency across different environments. Regarding Std, our method consistently exhibits lower values across all sequences, suggesting not only superior registration accuracy but also enhanced stability. This is critical for real-world applications where consistent performance is essential.

Overall, the average results indicate that our method achieves RMSE, Mean, and Std values of 3.08, 2.78, and 1.31, respectively, which significantly outperform the results of Fast-gicp (7.00, 5.68, 2.77), Faster-gicp (6.35, 5.59, 2.97), DLO (13.26, 12.49, 4.44), and Kiss-icp (4.04, 3.64, 1.74). Specifically, the average RMSE is reduced by at least 24\%, the Mean by at least 23\%, and the Std by at least 25\%, demonstrating the substantial advantages of our proposed method in terms of both registration accuracy and stability.

\begin{figure*}[htbp] 
	\centering 
	\includegraphics[width=0.85\linewidth]{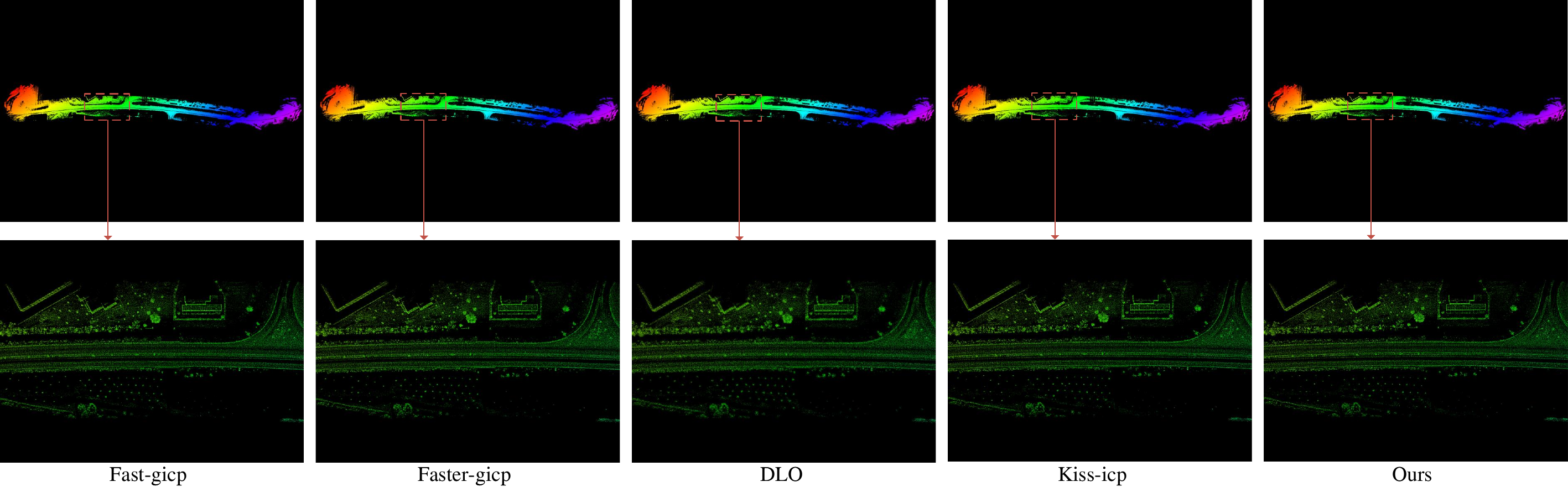}\\
	\caption{KITTI 01 global map comparison}
	\label{global01}
\end{figure*}

The experimental results clearly highlight the significant advantages of our proposed adaptive ICP-based LiDAR odometry method. Through reliable initial pose combined with adaptive registration, our method performs exceptionally well in complex dynamic environments, greatly improving the overall performance of LiDAR odometry. This validates the effectiveness of our method and provides strong support for advancing autonomous navigation and localization technologies for mobile robots.
\begin{figure*}[htbp] 
	\centering 
	\includegraphics[width=0.85\linewidth]{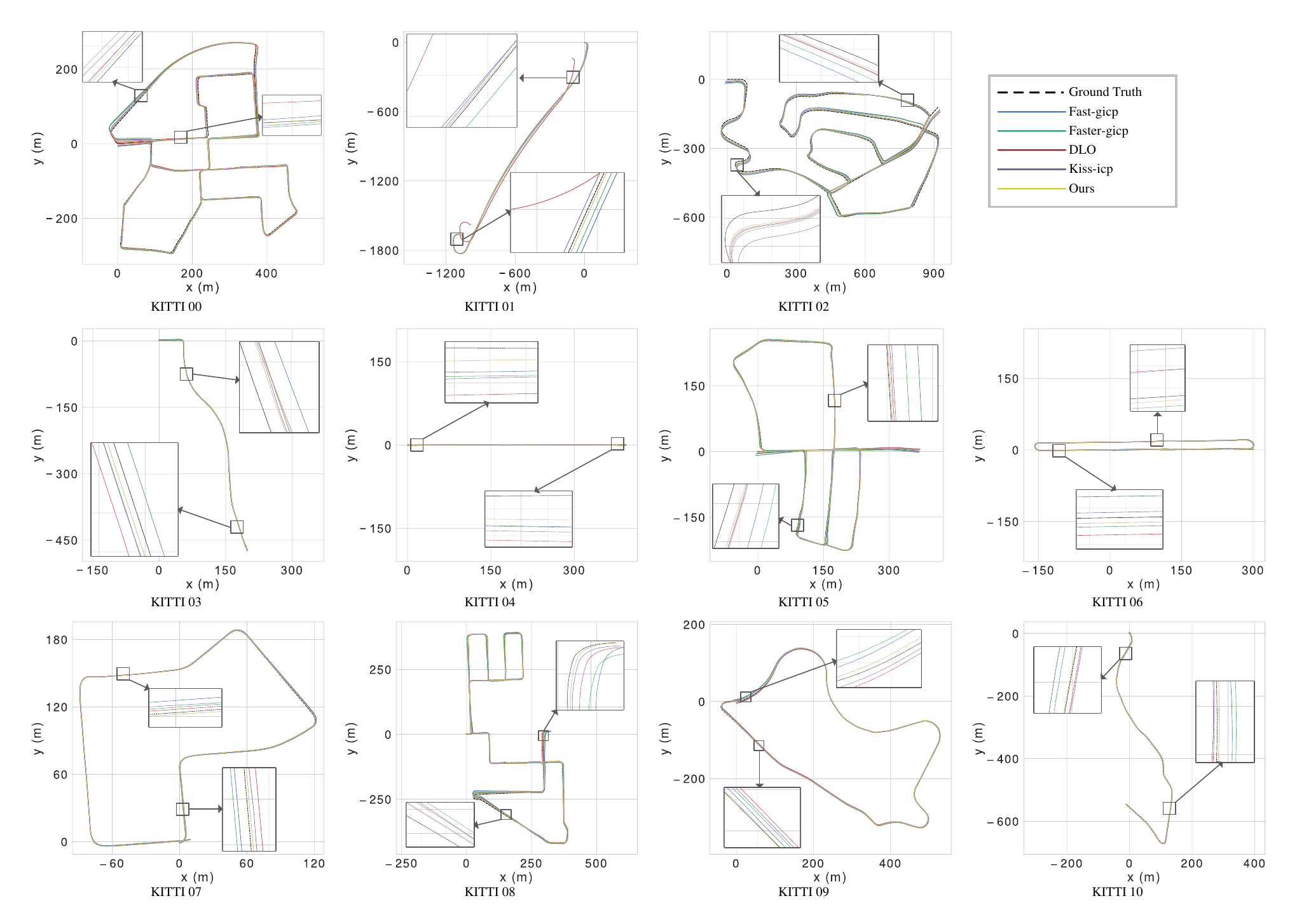}\\
	\caption{KITTI trajectory comparison}
	\label{02}
\end{figure*}

{In Fig. \ref{global01}, our method demonstrates the best mapping quality in high-speed scenarios compared to Fast-gicp, Faster-gicp, DLO, and Kiss-icp.  Our approach produces denser and more detailed point clouds, preserving structural integrity with minimal noise.  Unlike other methods, which exhibit sparse or blurry reconstructions, our method maintains sharp and continuous environmental features.  Moreover, it demonstrates strong robustness against motion-induced artifacts, ensuring a more reliable and accurate map.  These results highlight the superiority of our approach for high-precision mapping in dynamic environments.}
In Fig. \ref{02}, we compare the trajectories of Ours with other mainstream methods. It is evident that our method closely aligns with the ground truth trajectories across all sequences, exhibiting minimal deviation. Particularly in complex urban and highway scenarios, our method accurately captures the challenges posed by dynamic changes and high-speed movements, with trajectories nearly perfectly matching the real paths. In contrast, other methods, such as Fast-gicp and Faster-gicp, exhibit noticeable deviations in urban and highway environments, especially under high-density traffic and rapid turning conditions, leading to significantly increased errors. Similarly, in rural road sequences, our method maintains stable trajectory matching in open and less dynamic environments, while DLO and Kiss-icp perform adequately but still lag in overall accuracy compared to our approach.

{Our method achieves the lowest RMSE in most sequences; however, its performance in the city sequence 06 which contains dynamic objects is less impressive. This suggests that although the adaptive threshold effectively mitigates the influence of dynamic elements, a more robust approach may be required to further extend its lead. Additionally, a higher error is observed in sequence 01 (highway) where the LIDAR point cloud is relatively sparse due to high speed motion. This indicates that while our coarse registration with density filtering enhances initial pose estimation, occasional biases still occur in environments characterized by extremely sparse high speed motion. Overall, these results demonstrate the strengths of the proposed method while also highlighting potential areas for improvement such as the integration of additional sensors like Inertial Measurement Unit (IMU) or Global Positioning System (GPS) to further enhance robustness.}

{The experimental results presented in Table \ref{tab2} illustrate the average processing time per frame (in milliseconds) for different methods on the KITTI dataset (sequences 00-10). Our method demonstrates the lowest average processing time of 18.4 ms, significantly outperforming other approaches such as Fast-gicp (30.8 ms), Faster-gicp (27.2 ms), DLO (28.1 ms), and Kiss-icp (27.3 ms). This substantial improvement in efficiency is mainly due to two factors. First, our optimized initial pose estimation method, based on distributed coarse registration with density filtering, selects a reliable initial pose by comparing it with the motion prediction pose. This reduces the initial misalignment error, minimizing unnecessary iterations and avoiding excessive computational costs. Second, our adaptive frame-to-local map ICP registration dynamically adjusts registration parameters using both current and historical errors, allowing for more efficient convergence compared to traditional fixed-parameter approaches like Fast-gicp and DLO. This adaptive strategy optimizes the registration process in real-time, significantly reducing processing time while maintaining accuracy.
}

\begin{table}[]
	\centering
	\caption{ Comparison of processing time on KITTI (ms) }\label{tab2}
	\resizebox{0.8\linewidth}{!}{
	\setlength{\tabcolsep}{4pt} 
	\begin{tabular}{ccccccccccccc}
		\hline
		Sequence & 00  & 01  & 02  & 03  & 04  & 05  & 06  & 07  & 08  & 09  & 10  & Avg  \\ \hline
		Fast-gicp   & 40  & 22  & 37  & 27  & 24  & 35  & 21  & 39  & 28  & 30  & 36  & 30.8 \\
		Faster-gicp & 35  & 39  & \textbf{16}  & 38  & \textbf{15}  & 19  & 27  & 26  & 34  & 31  & 20  & 27.2 \\
		DLO         & 26  & 25  & 19  & 35  & 16  & 39  & 28  & 40  & 24  & 33  & 22  & 28.1 \\
		Kiss-icp    & 36  & 22  & 24  & 25  & 25  & 37  & 30  & 29  & 21  & 28  & \textbf{17}  & 27.3 \\
		Ours        & \textbf{18}  & \textbf{17}  & 18  & \textbf{23}  & \textbf{15}  & \textbf{17}  & \textbf{19}  & \textbf{20}  & \textbf{16}  & \textbf{21}  & 18  & \textbf{18.4} \\ \hline
	\end{tabular}
}
\end{table}

{The M2DGR dataset played an important role in evaluating the robustness of the system in challenging indoor scenarios, including scenes of randomly acting in halls (Hall01-Hall03) and rooms (Room01), entering and exiting lift (Lift01), and going from indoor to outdoor through a door (Door01).}

{The experimental results in Table \ref{tab3} demonstrate that our method achieves the lowest RMSE across all tested scenarios, showcasing superior robustness and accuracy. Compared to Fast-gicp, Faster-gicp, DLO, and Kiss-icp, our approach consistently maintains lower errors in challenging indoor environments such as Hall, Room, Door, and Lift scenarios. Notably, in the Lift01 scenario, our method achieves the best RMSE of 0.15, tied with Kiss-icp, while in Hall02 and Door01, our RMSE values of 0.25 and 0.28 significantly outperform other methods. This advantage primarily stems from two key improvements. First, we propose a distributed coarse registration method based on density filtering to estimate the initial pose. By selecting a reliable initial pose through comparison with the motion prediction pose, we effectively minimize the initial error between the source and target point clouds, thereby preventing the issue of local optimal solutions caused by unreliable initial poses. Second, we introduce an adaptive frame-to-local map ICP registration method, which dynamically adjusts registration parameters by integrating both current and historical errors. This enhances the adaptability of our approach in dynamic environments, making it particularly effective in scenarios with frequent environmental changes, such as Room01 and Door01. As a result, our method significantly outperforms others in handling complex point cloud registration tasks in dynamic indoor settings. Overall, the results on the M2DGR dataset validate the high accuracy and stability of our approach, particularly highlighting its strong adaptability and robustness in dynamic environments.}

\begin{figure}[htbp] 
	\centering 
	\includegraphics[width=0.85\linewidth]{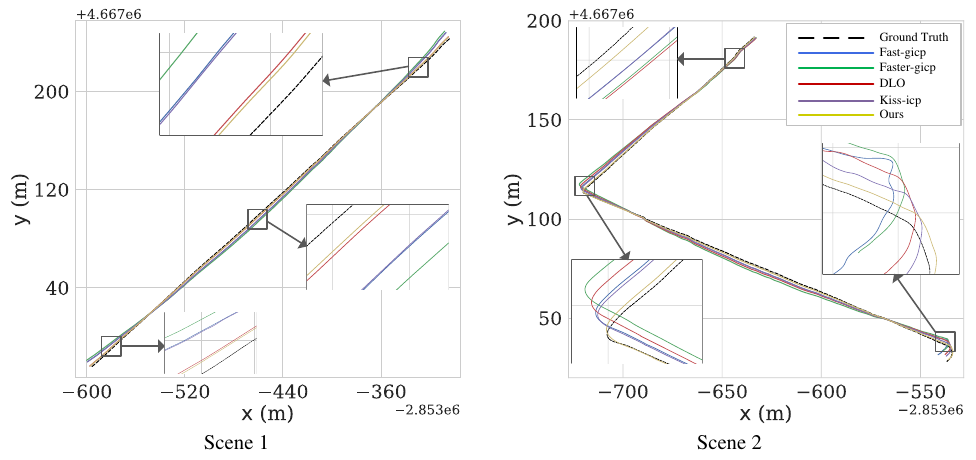}\\
	\caption{Real-world scenes trajectory comparison}
	\label{realworld}
\end{figure}

{In addition, we conduct comparative experiments between our method and the most mainstream methods in realistic outdoor scenes. From the Fig. \ref{realworld}, our method's trajectory shows minimal deviation from the ground truth, whereas other mainstream methods exhibit more noticeable discrepancies. In addition, as demonstrated by the APE comparison in Table \ref{tab4} for real-world scenarios, our method consistently achieves the lowest values across RMSE, Mean, and Std in both Scene 1 and Scene 2. For example, in Scene 1, our method's RMSE is only 0.83, while the RMSE values of the other methods are above 3.0; in Scene 2, our RMSE is 0.59, again significantly outperforming the alternatives. This indicates that in complex outdoor environments, our approach not only delivers higher accuracy but also exhibits enhanced stability and robustness.

}

\begin{table}[]
		\centering
	\caption{ RMSE comparison on M2DGR (METER) }\label{tab3}
	\resizebox{0.9\linewidth}{!}{
	\begin{tabular}{llllllll}
		\hline
		Sequnce     & Hall01        & Hall02        & Hall03        & Room01        & Door01        & Lift01        & Avg       \\ \hline
		Fast-gicp   & 0.52          & 0.41          & 0.45          & 0.54          & 0.47          & 0.36          & 0.46          \\
		Faster-gicp & 0.47          & 0.38          & 0.56          & 0.51          & 0.4           & 0.48          & 0.47          \\
		DLO         & 0.39          & 0.29          & 0.46          & 0.64          & 0.55          & 0.36          & 0.45          \\
		Kiss-icp    & 0.34          & 0.3           & 0.42          & 0.92          & 0.33          & \textbf{0.15} & 0.41          \\
		Ours        & \textbf{0.29} & \textbf{0.25} & \textbf{0.39} & \textbf{0.47} & \textbf{0.28} & \textbf{0.15} & \textbf{0.31} \\ \hline
	\end{tabular}
}
\end{table}

\begin{table}[]
	\centering
	\caption{ APE comparison of real-world scene (METER) }\label{tab4}
	\resizebox{0.9\linewidth}{!}{
\begin{tabular}{ccccccc}
	\hline
	\multirow{2}{*}{Scene} & \multicolumn{3}{c}{Scene   1}                 & \multicolumn{3}{c}{Scene   2}                \\ \cline{2-7} 
	& RMSE          & Mean          & Std           & RMSE          & Mean          & Std          \\ \hline
	Fast-gicp               & 3.09          & 2.68          & 1.53          & 2.29          & 1.93          & 1.23         \\
	Faster-gicp             & 4.07          & 3.55          & 1.98          & 3.53          & 3.06          & 1.77         \\
	DLO                     & 0.95          & 0.8           & 0.42          & 2.51          & 2.22          & 1.17         \\
	Kiss-icp                & 3.04          & 2.46          & 1.51          & 1.76          & 1.55          & 0.83         \\
	Ours                    & \textbf{0.83} & \textbf{0.72} & \textbf{0.36} & \textbf{0.59} & \textbf{0.55} & \textbf{0.2} \\ \hline
\end{tabular}
}
\end{table}

\begin{figure*}[htbp] 
		\centering 
		\includegraphics[width=0.75\linewidth]{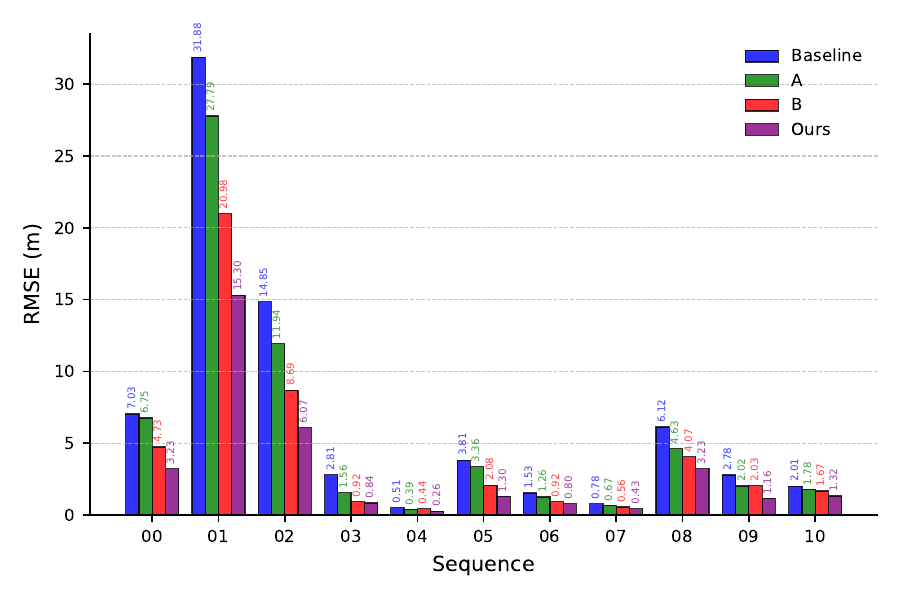}
		\caption{RMSE comparison for KITTI}
		\label{03}
\end{figure*}
	
\begin{figure}[htbp] 
	\centering 
	\includegraphics[width=0.81\linewidth]{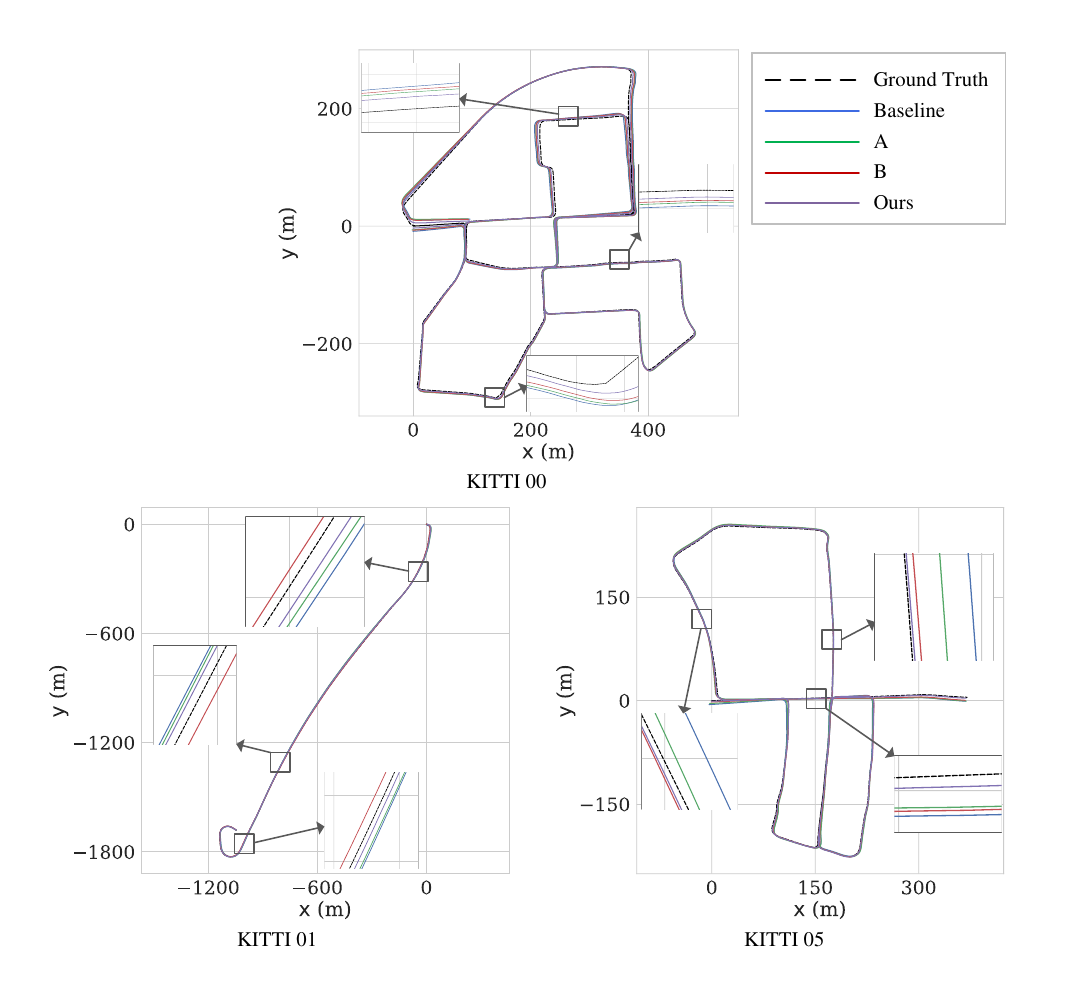}\\
	\caption{Trajectory comparison}
	\label{04}
\end{figure}
\subsection{Ablation Study}
We conducted ablation experiments on the KITTI dataset to evaluate the contributions of different components in our proposed method. In these experiments, the Baseline refers to the method without distributed coarse registration and adaptive ICP registration. Method A represents our approach without distributed coarse registration, while method B denotes our approach without adaptive ICP registration. These experiments aim to validate the effectiveness of each component of the proposed method.

As shown in Fig. \ref{03}, the Baseline method generally exhibits higher RMSE values across all sequences, with an average RMSE of 6.74, whereas Ours achieves an RMSE of 3.08. Method A, which excludes distributed coarse registration based on density filtering, has an average RMSE of 5.65, representing a reducion of approximately 16\% compared to the Baseline. This indicates the significance of adaptive ICP registration in enhancing registration accuracy and stability. Method B, which omits adaptive ICP registration from point cloud frames to the local map, achieves an average RMSE of 4.28, reducing the Baseline RMSE by about 36\%, thereby demonstrating the crucial role of distributed coarse registration based on density filtering. However, methods A and B exhibit RMSE values that are approximately 25\% and 16\% higher than Ours, respectively. This suggests that the synergy between both components is essential for achieving higher registration accuracy. Particularly in complex urban road and highway environments, Ours closely aligns with the ground truth trajectory, with RMSE values significantly lower than those of other methods.

{Fig. \ref{04} presents a trajectory-based comparison in three different scenarios (urban roads: sequence 00, highways: sequence 01, rural roads: sequence 05). In various scenarios, the Baseline method's trajectory appears more dispersed, indicating larger registration errors. Method A shows more concentrated trajectories after removing distributed coarse registration based on density filtering, but some deviations still exist. Method B further improves trajectory alignment, reducing deviations and enhancing accuracy. Ours demonstrates the closest alignment with the ground truth trajectories across all scenarios, highlighting its superior registration performance.}

In summary, the ablation experiments demonstrate that the combination of distributed coarse registration and adaptive ICP registration plays a crucial role in improving registration accuracy.  Distributed coarse registration provides a reliable initial pose, which helps stabilize the registration process, while adaptive ICP registration specifically addresses the challenges posed by dynamic environments, enhancing the accuracy of pose estimation.  The results show that the proposed method outperforms the individual components, highlighting the importance of their synergy in achieving optimal performance in complex, real-world driving environments.

{The sensitivity analysis in Table \ref{sensitivity_analysis} demonstrates how different values of key parameters affect the registration accuracy, measured by the average RMSE across KITTI sequences 00 to 10. The selected values in this study—density threshold percentile of 5.0, weight decay of 0.75, and decay factor of 1.5—achieve an RMSE of 3.08 meters, which is among the lowest across tested values, indicating a well-balanced parameter choice. A lower or higher density threshold tends to slightly increase RMSE, suggesting that an intermediate value effectively retains sufficient points for accurate registration while filtering out noise. Similarly, weight decay values that are too small or too large can degrade performance, likely due to improper handling of geometric error distribution. Lastly, the decay factor influences motion stability adaptation, and the chosen value of 1.5 provides an optimal trade-off between robustness and responsiveness to dynamic environments. This analysis confirms that the selected parameters ensure high accuracy and adaptability across different motion states and environmental conditions in KITTI sequences.
}

\begin{table}[]
	\centering
	\caption{Sensitivity Analysis of Key Parameters}
	\label{sensitivity_analysis}
	\resizebox{0.65\linewidth}{!}{
	\begin{tabular}{ccc}
		\hline
		\multicolumn{1}{c}{{Parameter}} & {Value} & {RMSE (m)} \\ \hline
		\multirow{5}{*}{Density Threshold percentile  $\alpha$} & 0  & 3.18  \\
		& 2.5  & 3.21  \\
		& 5.0  & 3.08  \\
		& 7.5  & 3.13  \\
		& 10.0  & 3.25 \\ \hline
		\multirow{5}{*}{Weight Decay $\sigma$} & 0.50  & 3.26  \\
		& 0.75  & 3.15  \\
		& 1.0  & 3.08  \\
		& 1.25  & 3.22  \\
		& 1.50  & 3.16  \\ \hline
		\multirow{5}{*}{Decay Factor $\sigma_{\mathrm{decay}}$} & 1.0  & 3.15  \\
		& 1.25  & 3.11  \\
		& 1.50  & 3.08  \\
		& 1.75  & 3.13  \\
		& 2.00  & 3.26  \\ \hline
	\end{tabular}
}
\end{table}

\section{Conclusion}
LiDAR odometry plays a critical role in autonomous navigation, especially in autonomous driving. Although ICP-based methods perform well in point cloud registration, traditional approaches often overlook the reliability of the initial pose, are susceptible to local optima, and lack adaptability, which significantly limits their accuracy in dynamic environments.

To address these challenges, this paper proposes an adaptive ICP-based LiDAR odometry method that leverages reliable initial pose estimates. Initially, the method obtains pose estimates through distributed coarse registration based on density filtering, followed by a comparison with motion-predicted poses to select the more reliable initial pose, thereby mitigating initial pose errors. Subsequently, adaptive thresholds are dynamically adjusted by combining current and historical errors, allowing the method to adapt to varying motion states and real-time dynamic environmental changes. Finally, with reliable initial values and adaptive thresholds, Point-to-Plane adaptive ICP registration is employed to achieve high-precision point cloud alignment.

Experimental results on the KITTI dataset show that our method outperforms conventional LiDAR odometry methods in registration accuracy and robustness. However, LiDAR-only odometry still faces challenges in highly dynamic, long-distance driving scenarios. Future work will focus on integrating visual sensors for multi-modal data fusion to further enhance localization accuracy and robustness in complex environments.

\bibliographystyle{IEEEtran}
\bibliography{ours.bib} 

\begin{IEEEbiography}
	[{\includegraphics[width=1in ,clip,keepaspectratio]{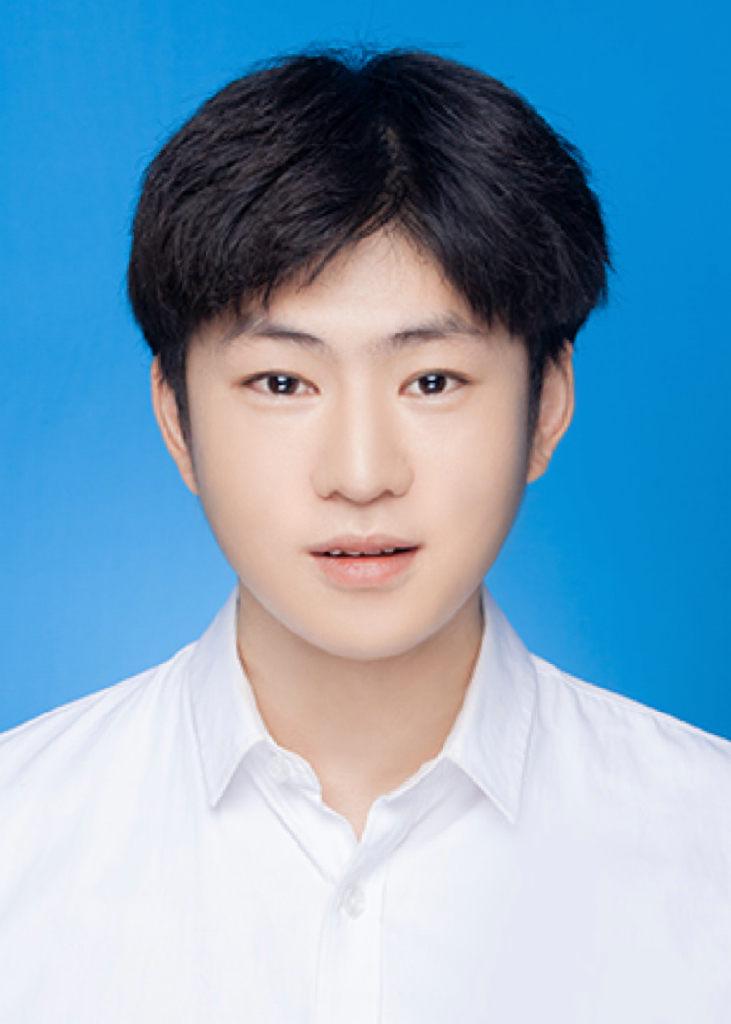}}] 
	{\bfseries Qifeng Wang} (Student Member, IEEE) received his master's degree from Wuhan University of Science and Technology in Wuhan, Hubei Province, in 2023. He is currently pursuing a PhD in Control Science and Engineering at Wuhan University of Science and Technology. His research interests include SLAM and point cloud registration.
\end{IEEEbiography}

\begin{IEEEbiography}
	[{\includegraphics[width=1in ,clip,keepaspectratio]{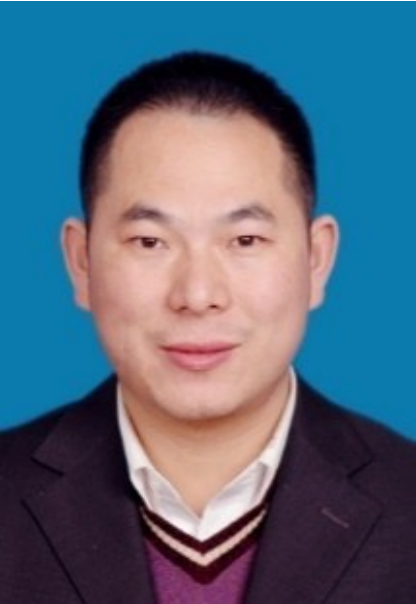}}] 
	{\bfseries Weigang Li} (Member, IEEE) is an professor and doctoral supervisor, as well as a specially appointed professor of Chu Tian Scholars, specializing in research on industrial artificial intelligence technology. He is the primary recipient of the first prize of the Hubei Province Science and Technology Progress Award. His research focuses on artificial intelligence and machine learning algorithms, as well as SLAM and 3D machine vision.
\end{IEEEbiography}

\begin{IEEEbiography}
	[{\includegraphics[width=1in ,clip,keepaspectratio]{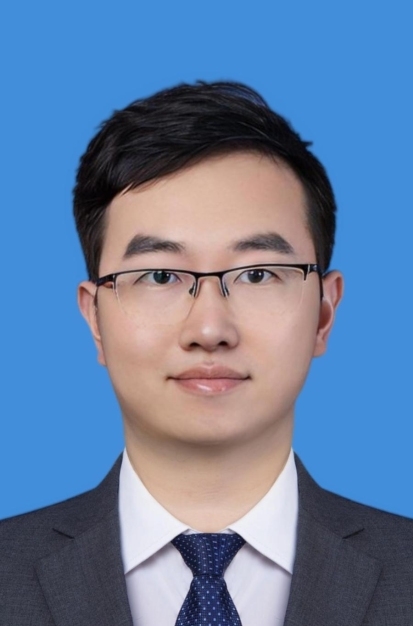}}] 
	{\bfseries Lei Nie} (Member, IEEE) received the B.S. degree in computer science and technology from Wuhan University of Science and Technology, Wuhan, China, in 2011, and the Ph.D. degree in computer architecture from Wuhan University, Wuhan, China, in 2017. He is currently a lecturer with the School of Computer Science and Technology, Wuhan University of Science and Technology. His main research interests include vehicular networks and intelligent transportation.
\end{IEEEbiography}

\begin{IEEEbiography}
	[{\includegraphics[width=1in ,clip,keepaspectratio]{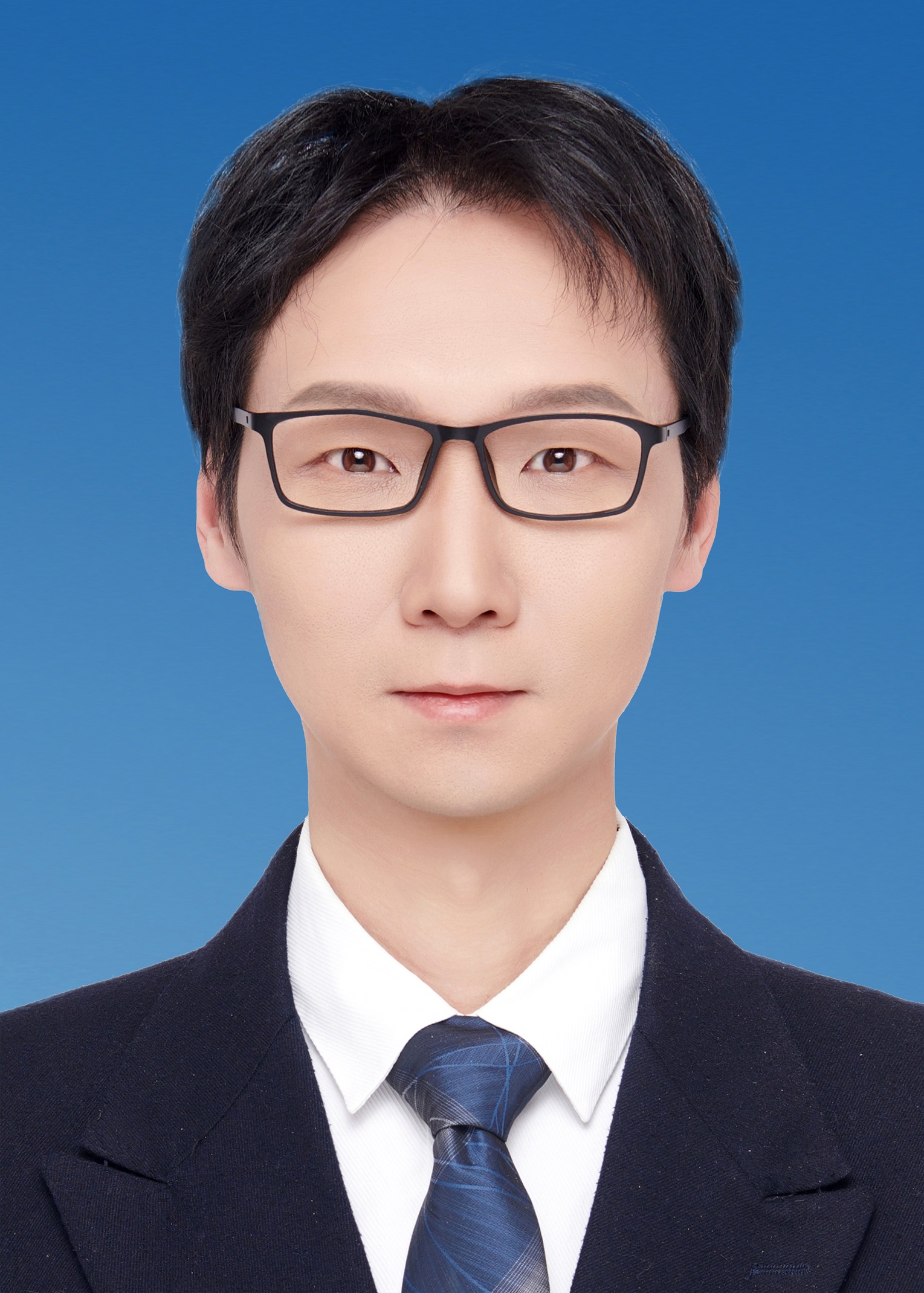}}]
	{\bfseries Xin Xu} (Senior Member, IEEE) received the B.S. and Ph.D. degrees in computer science and engineering from Shanghai Jiao Tong University, Shanghai, China, in 2004 and 2012, respectively. He is a Full Professor with the School of Computer Science and Technology, Wuhan University of Science and Technology, Wuhan, China. His research interests include artificial intelligence, computer vision, and image processing.
\end{IEEEbiography}

\begin{IEEEbiography}
	[{\includegraphics[width=1in ,clip,keepaspectratio]{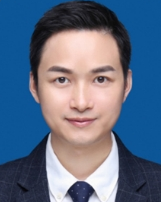}}] 
	{\bfseries Wenping Liu} (Senior Member, IEEE) is currently a Professor in Hubei University of Economics, China. He received the Ph.D. degree from Huazhong University of Science and Technology in 2012. His research interests include mobile computing, Internet of things, smart city and sensor networks.
\end{IEEEbiography}

\begin{IEEEbiography}[{\includegraphics[width=1in,clip,keepaspectratio]{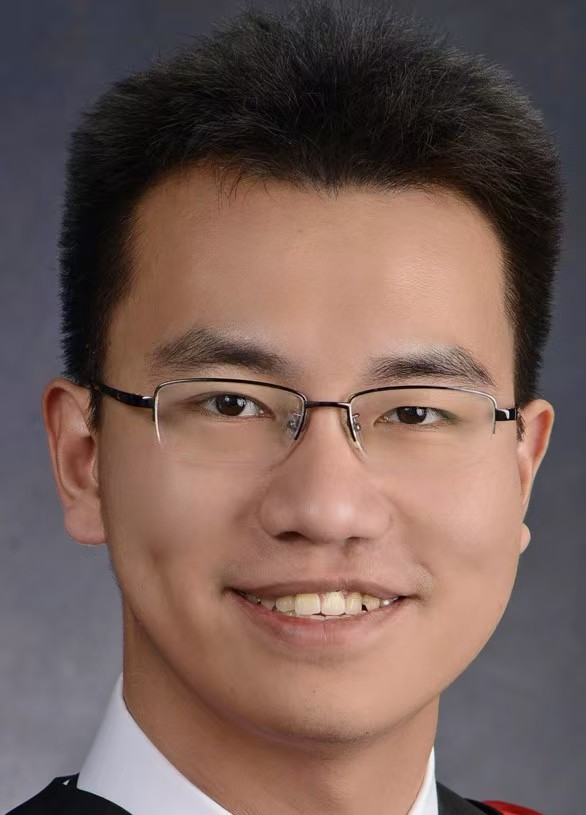}}]
	{\bfseries Zhe Xu} (Member, IEEE) received B.ENG. degree in Mechanical Engineering in 2018, and M.A.Sc. and Ph.D. degree in Engineering Systems and Computing in 2019 and 2023, respectively, from the University of Guelph. He is currently a post-doctoral fellow with the Department of Mechanical Engineering at McMaster University. His research interests include networked systems, tracking control, estimation theory, robotics, and intelligent systems.
\end{IEEEbiography}

\end{document}